\pdfoutput=1
\documentclass[runningheads]{llncs}

\usepackage[final,year=2026]{eccv}

\usepackage{eccvabbrv}

\usepackage{lmodern}

\usepackage{graphicx}
\usepackage{booktabs}
\usepackage{array}
\usepackage[protrusion=true,expansion=false]{microtype}
\usepackage{amsmath}
\usepackage{amssymb}
\usepackage{xcolor}
\usepackage{tikz}
\usetikzlibrary{arrows.meta,positioning,fit,backgrounds,calc}
\usepackage[accsupp]{axessibility}

\usepackage{hyperref}
\usepackage{orcidlink}
\usepackage{enumitem}

\begin{document}

\title{Depth-Wise Probing and Pruning of the Planning Token in a Driving Vision-Language-Action Model}

\titlerunning{Probing and Pruning a Driving VLA Planning Token}

\author{Harisankar Babu\inst{1,2}\orcidlink{0000-0003-0933-2936} \and
Benjamin Coors\inst{1} \and
Christopher Lang\inst{1}\orcidlink{0000-0002-8217-6485} \and
Hendrik Berkemeyer\inst{1} \and
Tamim Asfour\inst{2} \and
Simon Föll\inst{1}\orcidlink{0000-0002-4364-4282}
}
\authorrunning{H. Babu et al.}
\institute{Robert Bosch GmbH, Stuttgart, Germany \\
\email{\{harisankar.babu,simon.foell\}@bosch.com}
\and Karlsruhe Institute of Technology, Karlsruhe, Germany}

\maketitle

\begin{abstract}
Vision-language-action (VLA) models route driving decisions through a deep
language model, but it is unclear how much of that depth the action itself
requires. We study a representative driving VLA whose entire plan is carried by
a single planning token that a generative planner decodes into a
trajectory. Borrowing the planner as a trajectory-space logit lens, we decode
the planning token from every one of the 32 decoder layers and measure two
signals: the linear decodability of the navigation command and trajectory
compatibility with the frozen native planner. Our diagnostic shows that semantic
intent is linearly decodable early: command-probe accuracy reaches 97.7\% after
the first decoder layer, compared with 16.7\% chance. In contrast, compatibility with
the frozen native planner improves gradually across depth, with open-loop
Avg-L2 reaching its minimum of 2.11\,m only at the final layer. Learned
readouts from the first layer recover much of this gap, indicating that planning
information is already present early but is not yet represented in the format
expected by the deployed planner. Ranking decoder layers by the angular
deviation they induce in the planning token permits removal of 8 of 32 layers
within an approximately 5\% relative open-loop error increase and yields a
measured 1.33$\times$ decoder speedup. At the evaluated sample size, no
family-specific degradation is statistically resolved. These findings are
limited to the evaluated ORION checkpoint and Bench2Drive setup.
\keywords{Vision-language-action models \and Autonomous driving \and
Interpretability \and Layer pruning}
\end{abstract}

\section{Introduction}
\label{sec:intro}

End-to-end driving models increasingly place a large language model at the core
of the computational stack. In this paradigm, perception outputs are projected
into tokens, a transformer architecture reasons over the sequence, and an action
head projects the resulting representations into a trajectory~\cite{fu2025orion,tian2024drivevlm,shao2024lmdrive}.
While these architectures leverage broad world knowledge to handle rare or
unstructured driving scenarios, this structural depth introduces substantial
computational latency that safety-critical controllers cannot afford. Rather than
relying solely on post-hoc acceleration or distillation, optimizing these pipelines
requires an empirical investigation into the spatial distribution of task-relevant
computation across the network layers. Specifically, it remains unclear how
planner-relevant computation is distributed across decoder depth and whether some
layers can be removed without substantial open-loop degradation.

We study this question in the representative vision-language-action (VLA) design paradigm
(as implemented in architectures such as ORION~\cite{fu2025orion}) whose structure makes
this question sharp under a trajectory-space logit lens. The model compresses its entire plan
into a single hidden state, called the planning token, that a generative planner decodes
into future waypoints. Because this vector is the sole interface between language reasoning
and action, we can ask exactly when, along the decoder, the information the planner
needs becomes available.

Our method reads the planning token out of every intermediate layer and decodes it with the
model's native planner, scoring the resulting trajectory. This is a
trajectory-space analogue of the logit lens~\cite{nostalgebraist2020logitlens,belrose2023tunedlens}.
Rather than measuring what a layer could encode with a trained readout, we ask what
the model's own downstream module makes of each layer. Alongside the trajectory,
we read the discrete navigation intent with a linear probe~\cite{alain2017probes}.

Our empirical findings indicate that semantic intent and native-planner
compatibility follow different depth-wise schedules. The navigation command is
linearly decodable from the first decoder block and remains highly decodable
throughout the network, despite the command never being provided to the model as
textual input. By contrast, the same intermediate planning token becomes
progressively more compatible with the frozen native planner, reaching its
lowest trajectory error only at the final hidden state. Learned layer-1 readouts
recover much of the gap, showing that the frozen-planner curve should be
interpreted as a measure of representation compatibility rather than the first
presence of geometric information. We extend this analysis across the five
Bench2Drive ability categories~\cite{jia2024bench2drive}, where the progression
of planner compatibility is consistent.

The depth-wise diagnostic also motivates a pruning test: layers that minimally
rotate the planning token may be removable even when their isolated
native-planner readout is poor. We rank layers by the angular deviation they
induce in the planning token, prune the decoder to trace an accuracy-latency
frontier paired with the measured wall-clock of a physically shortened decoder,
and break the error down by ability category. At the evaluated sample size, no
family-specific degradation is statistically resolved.

Our contributions are:
\begin{itemize}[nosep]
\item A trajectory-space native-head lens for measuring how intermediate
planning-token representations become compatible with the deployed frozen
planner, alongside a linear probe of navigation intent.
\item A depth-wise analysis across five Bench2Drive ability categories, showing
early linear command decodability and a consistent progression of planner
compatibility across the evaluated ORION checkpoint.
\item A pruning study showing that planning-token input-output cosine is a better
removal criterion than isolated intermediate decode quality: 8 of 32 layers can
be removed within an approximately 5\% relative open-loop error increase,
yielding a measured 1.33$\times$ decoder speedup.
\end{itemize}

\section{Related Work}
\label{sec:related}

\noindent\textbf{Language and VLA Models for Driving.}
Recent systems integrate large language or vision-language models into the driving stack.
One line uses the language model as a symbolic reasoner that emits discrete textual
plans~\cite{mao2023gptdriver,mao2024agentdriver,sima2024drivelm,nie2024reason2drive,wang2023drivemlm};
another emits continuous trajectories or control tokens end-to-end via fast-slow~\cite{tian2024drivevlm}
or unified action networks~\cite{shao2024lmdrive,xu2024drivegpt4,jiang2024senna,hwang2024emma,fu2025orion},
paralleling instruction-to-action robotics~\cite{zitkovich2023rt2,kim2024openvla}. These inherit the
latency of LLaMA-scale backbones~\cite{touvron2023llama,vaswani2017attention}. We analyze the spatial
distribution of trajectory computation within such stacks, using ORION~\cite{fu2025orion} as a
representative model, rather than proposing a new architecture.

\noindent\textbf{End-to-End Driving and Representation Readout.}
Earlier end-to-end networks fused sensors to predict trajectories~\cite{prakash2021multimodal,chitta2023transfuser,hu2022stp3}
or unified perception, prediction, and planning via query-based systems~\cite{hu2023uniad,jiang2023vad}
and world models~\cite{zheng2024genad,yang2024genad}. We evaluate on the CARLA-based Bench2Drive
benchmark~\cite{dosovitskiy2017carla,jia2024bench2drive}, scoring open-loop trajectories with ST-P3
ADE~\cite{hu2022stp3}. Linear probes read features accessible to an external classifier~\cite{alain2017probes},
whereas the logit and tuned lens decode intermediate layers through the model's native
heads~\cite{nostalgebraist2020logitlens,belrose2023tunedlens}; we extend the latter to continuous
action by using the frozen planner as the output head.

\noindent\textbf{Depth Reduction in Transformers.}
Transformers often exhibit layer redundancy, addressed by structured dropout~\cite{fan2020layerdrop},
post-hoc deletion~\cite{sajjad2023dropping}, or similarity-guided pruning~\cite{men2025shortgpt,gromov2025unreasonable,ma2023llmpruner,ashkboos2024slicegpt},
and by dynamic early exit~\cite{teerapittayanon2016branchynet,schuster2022confident,elhoushi2024layerskip}.
Our contribution is the trajectory-space lens itself; we use pruning to validate the diagnostic
rather than to propose a compression method, and we report pruning effects across driving abilities.

\section{Background: the Model and its Planning Token}
\label{sec:background}

\begin{figure*}[tp]
  \centering
  \resizebox{\textwidth}{!}{%
  \begin{tikzpicture}[
    font=\footnotesize,
    box/.style={draw, rounded corners=2pt, align=center, minimum height=8.5mm,
                inner xsep=4pt, inner ysep=3pt, line width=0.5pt},
    vis/.style={box, fill=blue!7},
    llm/.style={box, fill=orange!12, minimum height=15mm},
    plan/.style={box, fill=red!12, line width=0.9pt},
    plng/.style={box, fill=green!9},
    tok/.style={box, fill=black!5},
    arr/.style={-{Latex[length=2mm]}, line width=0.6pt},
    darr/.style={-{Latex[length=1.8mm]}, line width=0.6pt, dashed, gray!70!black},
  ]
    \node[vis] (enc) {vision backbone\\+ perception heads};
    \node[tok, right=5mm of enc] (ptok) {perception\\tokens};
    \node[tok, below=3mm of ptok] (prompt) {fixed text\\prompt};
    \node[llm, right=6mm of ptok] (llm) {32-layer\\LLM decoder};
    \node[plan, right=6mm of llm] (plan) {planning\\token};
    \node[plng, right=6mm of plan] (planner) {frozen\\planner};
    \node[tok, right=5mm of planner] (wp) {6 waypoints\\(3\,s)};
    \node[left=5mm of enc, align=center] (cam) {6 surround\\cameras};

    \draw[arr] (cam) -- (enc);
    \draw[arr] (enc) -- (ptok);
    \draw[arr] (ptok.east) -- ([yshift=3mm]llm.west);
    \draw[arr] (prompt.east) -- ([yshift=-3mm]llm.west);
    \draw[arr] (llm) -- (plan);
    \draw[arr] (plan) -- (planner);
    \draw[arr] (planner) -- (wp);

    \node[plng, below=9mm of planner, minimum height=7mm] (pprobe)
      {same frozen\\planner};
    \node[tok, right=5mm of pprobe, minimum height=7mm] (ptraj) {trajectory\\at layer $\ell$};
    \draw[darr] (llm.south) |- (pprobe.west)
      node[midway, below, xshift=-6mm, text=gray!70!black] {read token at layer $\ell$};
    \draw[darr] (pprobe) -- (ptraj);
    \node[align=center, font=\scriptsize\itshape, text=gray!55!black,
          below=1mm of pprobe.south west, anchor=north west, xshift=-14mm]
      {our probe};

    \begin{scope}[on background layer]
      \node[fit=(enc)(ptok)(prompt), rounded corners, draw=blue!30, line width=0.4pt,
            inner sep=3pt] {};
      \node[fit=(planner)(wp), rounded corners, draw=green!35, line width=0.4pt,
            inner sep=3pt] {};
    \end{scope}
  \end{tikzpicture}%
  }
  \caption{The driving VLA and our probe. The decoder receives perception tokens
  (object and map queries projected into the token space) and a fixed text
  prompt; the navigation command is never given as text. The model appends one
  special token whose final-layer hidden state, the planning token, is
  decoded into a trajectory by a frozen generative planner. Our probe reads that
  token from \emph{every} layer, decodes it with the same frozen planner, and
  scores it against the ground truth.}
  \label{fig:pipeline}
\end{figure*}
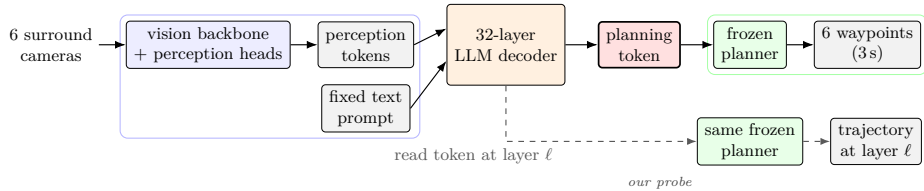

The evaluated model, ORION~\cite{fu2025orion}, processes six surround-view camera streams through
a vision backbone and a temporal QT-Former coupled with two specialized perception heads tracking
dynamic objects and localized map features. Grounded in Bench2Drive and Chat-B2D sequence training,
these heads emit several hundred query tokens that are projected into the language model's embedding
space and concatenated with a static text prompt. While some VLA variants include dynamic features like
ego-vehicle velocity or prior waypoints within the textual input, the high-level routing command itself
is never provided to the model as textual instructions. The model appends a single special token and runs
a 32-layer LLaMA-style decoder. The 4096-dimensional vector at this special token position, extracted
after the final layer, constitutes the planning token (\cref{fig:pipeline}).

A compact generative planner, trained jointly with the base model and kept frozen throughout
this study, maps this planning token to continuous coordinates. The planner consists of a
conditional variational autoencoder (VAE) that transforms the planning token into a latent
vector, a recurrent network that projects this latent vector into sequential future states,
and a final decoder head. This head outputs candidate displacement vectors for each driving
mode across a 3\,s horizon divided into six time steps. At evaluation time, the ground-truth
one-hot navigation command selects the retained trajectory mode; this selector is held fixed
across all intermediate hidden states, is not provided to the language-model input, and is not
injected into the planning-token representation. The retained mode's displacements are
cumulatively summed to establish absolute metric waypoints. Throughout, the planner is used
exactly as released; we never retrain it.

Because the command is only a loss weight and a downstream mode selector, never a text
input, any command-specific signal in the intermediate planning token must be generated
internally by the model, plausibly from the predicted objects and map boundaries derived
from the vision tokens. The model therefore cannot copy the command from text or read it
off historical ego-trajectories.

\section{Probing the Planning Token}
\label{sec:method}

\noindent\textbf{Probing by decoding every layer.}
A standard probe trains a classifier on a layer's activations and evaluates what could
be read out. We instead evaluate how the model's \emph{own} planner interprets the
representations at each layer. At each of the 33 hidden states (the embedding output and the output of
each of the 32 layers), we take the planning token, apply the model's final
normalization (the same one the planner sees in normal operation), and pass it
through the frozen planner to produce a trajectory. Reading layer $\ell$ this way
answers a counterfactual: if the model had to commit to a plan using its
layer-$\ell$ representation, how good would the plan be? Because the planner is
fixed, every layer is evaluated under an identical metric space. Accordingly, this
curve measures compatibility with the deployed frozen planner, not whether geometric
information could be recovered by a separately trained readout. This is analogous to
the logit lens~\cite{nostalgebraist2020logitlens,belrose2023tunedlens} applied to
trajectories instead of to vocabulary logits.

\noindent\textbf{Trajectory evaluation metric.}
We score each decoded trajectory against the ground truth with the ST-P3 Average Displacement Error
(ADE) displacement metric~\cite{hu2022stp3}. Let $\hat{p}_t$ and $p_t$ be the predicted
and ground-truth positions at waypoint $t$. The cumulative average displacement error (ADE)
at horizon $H$ is the average Euclidean distance up to $H$, $\frac{1}{H}\sum_{t=1}^{H}\lVert\hat{p}_t - p_t\rVert_2$.
Predicted and ground-truth displacements are cumulatively summed into ego-frame waypoints before
scoring. We report Avg-L2 over all six waypoints (equivalently plan-L2 at 3\,s), L2@2s over the
first four, and Final-L2 at the sixth waypoint. Lower is better; all values are in meters, with 95\%
confidence intervals over sampled frames.

\noindent\textbf{Semantic command probing.}
At each hidden state we fit an $\ell_2$-regularized multinomial logistic regression classifier to
predict the six Bench2Drive commands. We report frame-level stratified five-fold cross-validation
accuracy with balanced class weights. The folds are not grouped by route or scene, and the feature
standardizer is fitted before fold construction. The result should therefore be interpreted as
frame-level linear decodability and may benefit from route correlations and minor normalization
leakage. Chance is $1/6\approx 17\%$. Because the command is never an input, this probe measures the
linear separability of the learned feature representations for distinct driving commands.

\noindent\textbf{Deterministic inference configuration.}
The generative planner is variational, but operates deterministically at inference by
decoding at the latent mean, reflecting the model's native deployment configuration.
Empirically, the released checkpoint exhibits posterior collapse: across eight
independent latent draws per scene, the predicted trajectory shifts by less than
0.001\,m, making the choice of latent sampling immaterial.

\noindent\textbf{Ability category taxonomy and set assignments.}
Every validation frame belongs to a Bench2Drive scenario type, which we read from the scene
metadata. We map each scenario to one of the five official Bench2Drive ability categories~\cite{jia2024bench2drive}
(merging, overtaking, emergency braking, giving way, and traffic-sign compliance). Because
these original evaluation sets overlap, we partition each scenario to a single primary category
via a fixed rule detailed alongside exact scenario frame counts in \cref{tab:membership}.

\begin{table}[!ht]
  \centering
  \caption{Primary capability assignment of Bench2Drive scenario types, with frame
  counts in the validation split. Parenthetical suffixes denote variants (for
  example \emph{Accident(TwoWays)} covers Accident and AccidentTwoWays).}
  \label{tab:membership}
  \setlength{\tabcolsep}{6pt}
  \renewcommand{\arraystretch}{1.1}
  \small
  \begin{tabular}{@{}p{0.18\textwidth}>{\raggedright\arraybackslash}p{0.68\textwidth}r@{}}
    \toprule
    Capability & Scenario types & Frames \\
    \midrule
    Emergency brake & DynamicObjectCrossing, PedestrianCrossing,
      ParkingCrossingPedestrian, VehicleTurningRoute(Pedestrian), HardBreakRoute,
      StaticCutIn, ParkingCutIn, ControlLoss, OppositeVehicleTakingPriority,
      OppositeVehicleRunningRedLight, BlockedIntersection & 3754 \\
    \addlinespace[2pt]
    Traffic sign & SignalizedJunctionLeft/RightTurn(EnterFlow), TJunction,
      VanillaSignalizedTurnEncounterRed/GreenLight,
      VanillaNonSignalizedTurnEncounterStopsign & 3260 \\
    \addlinespace[2pt]
    Overtaking & Accident(TwoWays), ConstructionObstacle(TwoWays),
      ParkedObstacle(TwoWays), HazardAtSideLane(TwoWays), VehicleOpensDoorTwoWays
      & 2864 \\
    \addlinespace[2pt]
    Merging & EnterActorFlow, CrossingBicycleFlow, HighwayExit, HighwayCutIn,
      Interurban(Advanced)ActorFlow, MergerIntoSlowTraffic(V2),
      NonSignalizedJunctionLeft/RightTurn(EnterFlow), ParkingExit, LaneChange
      & 2458 \\
    \addlinespace[2pt]
    Give way & YieldToEmergencyVehicle, InvadingTurn & 470 \\
    \bottomrule
  \end{tabular}
\end{table}

\noindent\textbf{Layer pruning formulation.}
To test whether depth is needed, we remove decoder layers at inference. Each
LLaMA layer adds a pure residual, so patching layer $k$ to the identity is
equivalent to deleting it from the stack; we confirm that the identity patch and
a physically sliced decoder produce the same planning token (cosine $1.0$; \cref{sec:prune}).
We rank layers for removal by the angular deviation they induce in the planning token
(the cosine similarity between a layer's input and output hidden states, where a high cosine
denotes a layer that minimally rotates the planning token) and prune the highest-cosine layers
first; this metric is invariant to magnitude scaling. We compare this representational order
against a decode-marginal order, a contiguous late-block order, and a random order.

\noindent\textbf{Computational latency metrics.}
The layers-executed fraction $(32-k)/32$ is a hardware-independent proxy. We also measure the
wall-clock of one non-autoregressive FP32 decoder forward pass at batch size 1 on a single
A100 80\,GB GPU, averaged over many runs after warm-up; this measurement excludes language
generation and planner decoding.

\noindent\textbf{Evaluation protocol.}
We use four evaluation sets. The probing set contains 2{,}000 ability-balanced cold-start frames;
the pruning set contains 800 ability-balanced cold-start frames; the reshaping set contains 500
command-balanced cold-start frames; and the streaming set contains 5{,}291 route-ordered frames.
Cold-start resets the temporal memory of both perception heads for every frame. Streaming processes
complete routes in temporal order and retains perception memory until a route boundary. Results from
different sets should not be compared at centimeter scale. On the same 5{,}291 route frames,
streaming lowers the 1, 2, and 3\,s errors by 0.008, 0.035, and 0.073\,m while preserving the
depth-wise trend. Absolute errors are higher than published full-validation numbers because we
re-weight the scene split distribution.

\section{Where Intent and Planner Compatibility Develop}
\label{sec:emerge}

\begin{figure}[tbp]
  \centering
  \includegraphics[width=0.66\textwidth]{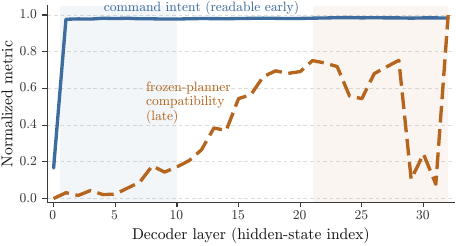}
  \caption{Command intent is readable early; planner compatibility is refined late.
  Command accuracy is 16.7\% at hidden state~0, 97.7\% at hidden state~1, and peaks
  at 98.6\%. The lower curve shows normalized frozen-planner Avg-L2 and should not be
  interpreted as using a symmetric readout. Cold-start, ability-balanced probing set.}
  \label{fig:semgeo}
\end{figure}

The planning token carries two kinds of information that become available at
different depths (\cref{fig:semgeo}). The command probe is at chance at hidden
state 0, where the appended planning token is still context-free. Accuracy rises
to 97.7\% after the first decoder block and peaks at 98.6\%. Because the command
is not provided in the prompt, this shows that command-related structure becomes
linearly accessible after interaction with the multimodal context.

\begin{figure}[tbp]
  \centering
  \includegraphics[width=\textwidth]{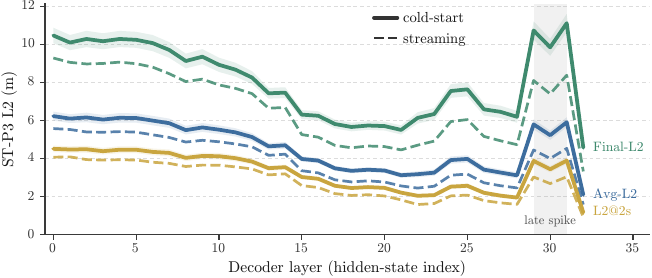}
  \caption{Per-layer trajectory error for the three ST-P3 metrics, cold-start
  (solid, with 95\% CI bands) and streaming (dashed). Error falls with depth,
  spikes at hidden-state indices 29 to 31, and is lowest at the final hidden state.
  The streaming curve is lower but identically shaped.}
  \label{fig:layercurve}
\end{figure}

Compatibility with the frozen planner follows a different schedule. Error falls
slowly through the middle layers and reaches its minimum only at the final hidden
state, where Avg-L2 is 2.11\,m; more than half of the total reduction over the
embedding baseline is reached only by hidden state~15, as expanded across metric
configurations in \cref{fig:layercurve}. Early command decodability does not imply
early compatibility with the frozen planner: at hidden state~1 the command probe is
already near its ceiling, yet decoding the same token gives an Avg-L2 worse than the
embedding baseline. The early layers hold the discrete intent in a form a linear classifier
can read, but not yet in a form the frozen planner can turn into a low-error
trajectory. The representation becomes progressively more compatible with the
frozen planner and reaches its lowest error only at the final hidden state. The
continuous per-layer trends (\cref{fig:layercurve}) show a late non-monotonic
spike in native-planner error at hidden-state indices 29 to 31 before the final
hidden state recovers. We treat this as an observed directional excursion rather
than a resolved mechanism, and we return to it when we prune.

\section{Planner Compatibility Across Scenarios}
\label{sec:edgecase}

\begin{figure*}[tp]
  \centering
  \includegraphics[width=\textwidth]{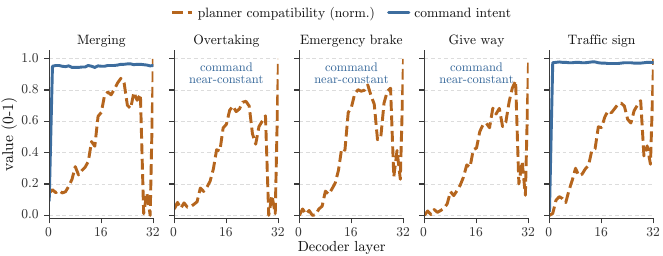}
  \caption{Per-family planner compatibility across ability categories. For merging and
  traffic-sign scenarios, where the navigation command has sufficient variation, it
  is linearly decodable after the first block (blue). The command probe is omitted
  for emergency-brake, give-way, and overtaking subsets because their command
  distributions are near-constant or strongly imbalanced. The dashed curve reports
  normalized compatibility with the frozen planner, which rises late in every family
  at a consistent depth.}
  \label{fig:ability}
\end{figure*}

The separation identified in \cref{sec:emerge} could reflect an artifact of
averaging over routine frames. To evaluate across diverse driving scenarios,
\cref{fig:ability} stratifies the analysis for each Bench2Drive ability category,
and \cref{tab:emergence} summarizes the depths. The normalized trajectory quality
in \cref{fig:semgeo} scales the global Avg-L2 to $[0,1]$, with an absolute minimum
of $2.11$\,m (hidden state~32) and an absolute maximum of $6.23$\,m (hidden state~0).

This stratification reveals two trends. First, the planner compatibility
progression is consistent across ability categories: half of the total error
reduction is reached by hidden state~15 in four of the five families (hidden
state~17 for give-way), with the minimum at the final hidden state in all five.
Second, where the navigation command has sufficient variation (merging and
traffic-sign compliance), it is linearly readable from the first hidden state
(95 to 98\% at hidden state~1). In the emergency-brake, give-way, and overtaking
families the command distribution is near-constant or strongly imbalanced, so the
linear command probe is uninformative there. The property that holds across every
ability category is the consistent, late depth at which planner compatibility
develops.

\begin{table}[t]
  \centering
  \caption{Intent and compatibility depth by capability (hidden-state index, 0 to 32) on the
  ability-balanced probing set. ``Intent'' is the hidden state at which the command
  probe reaches 90\% of its peak; it is undefined (n/a) where the command is
  near-constant or strongly imbalanced. ``Compatibility 50\%'' is the hidden state at
  which trajectory error reaches half of its total reduction under the frozen-planner
  readout; the minimum occurs at the final hidden state for every family.}
  \label{tab:emergence}
  \begin{tabular}{lccc}
    \toprule
    Capability & $n$ & Intent & Compatibility 50\% \\
    \midrule
    Merging          & 400 & 1               & 15 \\
    Overtaking       & 400 & n/a$^{\dagger}$ & 15 \\
    Emergency brake  & 400 & n/a             & 15 \\
    Give way         & 400 & n/a             & 17 \\
    Traffic sign     & 400 & 1               & 15 \\
    \bottomrule
  \end{tabular}
  \\[3pt]
  {\footnotesize $^{\dagger}$Overtaking is approximately 99\% one command in this
  subset, so its hidden-state-0 majority-class accuracy satisfies the 90\%-of-peak
  threshold; we treat the intent depth as undefined rather than meaningful.}
\end{table}

\section{How Much Depth Can We Prune?}
\label{sec:prune}

\begin{figure}[tbp]
  \centering
  \includegraphics[width=\textwidth]{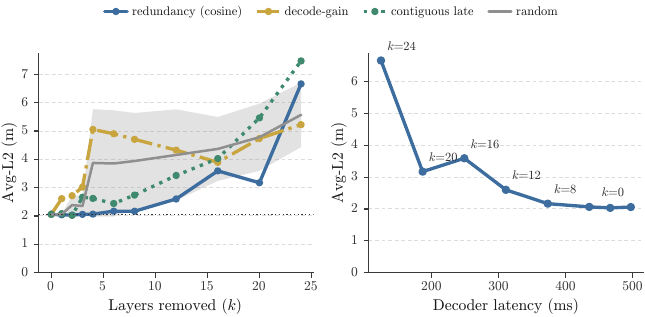}
  \caption{Depth-pruning frontier. Left: Avg-L2 versus the number of removed
  layers for the representational-redundancy order (highest input-output cosine
  first), a decode-marginal order, a contiguous late-block order, and a random
  order. Right: error versus measured decoder latency; markers give the
  layers-executed fraction.}
  \label{fig:pareto}
\end{figure}

Layers that minimally rotate the planning token may be removable even when their
isolated frozen-planner readout is poor. \Cref{fig:pareto} traces the open-loop error as we
remove the $k$ layers with the highest input-output cosine first. The ranking is
computed once from input-output cosine at the planning-token position, averaged
over scenes, and is fixed across all pruning and latency experiments. Error stays
within $\sim$5\% of the full model up to $k=8$ removed layers, and within 1\% up to
$k=4$ (Avg-L2 $2.06\to2.17$\,m; the first removals leave it unchanged), then rises
sharply once the prune set reaches a computationally critical layer. Using
zero-based decoder-layer indices, the removed set at $k=8$ is
$\{9,10,13,17,18,19,27,30\}$. Removing these eight layers reduces decoder latency
from 497.52 to 373.22\,ms, a measured 1.33$\times$ speedup (layers-executed fraction
$0.75$); larger removals trade more error for more speed (up to $4\times$ at $k=24$).
The identity-patch proxy and a physically sliced decoder produce the same planning
token (cosine $1.0$ at every $k$), so the error and latency numbers describe the
same shortened model.

\begin{figure}[tbp]
  \centering
  \includegraphics[width=\textwidth]{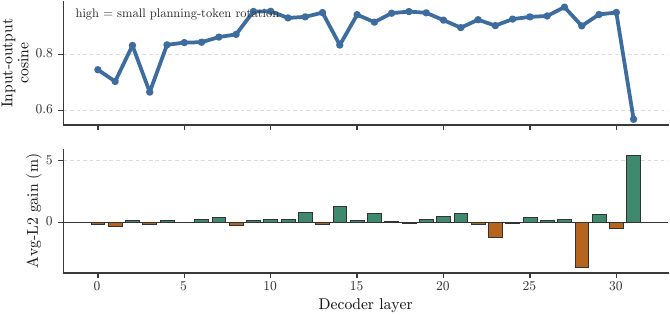}
  \caption{Planning-token redundancy diagnostics. Top: input-output cosine at the
  planning-token position, averaged over scenes. High cosine indicates a small
  directional update, not that the complete decoder layer is globally unimportant.
  Bottom: isolated frozen-planner Avg-L2 gain. Hidden-state indices 29 to 31 exhibit
  the late decode spike.}
  \label{fig:redundancy}
\end{figure}

The prune \emph{order} matters. Ranking layers by their own decoded error (removing
the layers that individually decode worst) is a poor guide: that order targets the
late spike (hidden-state indices 29 to 31), which decode badly in isolation yet are
needed by the layers after them, and it raises error by 27\% at the first layer
removed. Ranking instead by the angular deviation they induce (\cref{fig:redundancy},
top) primarily selects middle and upper-middle layers with high planning-token cosine
and stays within $\sim$5\% up to $k=8$ (\cref{fig:pareto}, left). As illustrated by
the redundancy diagnostics in \cref{fig:redundancy}, a contiguous late-block order
and a random order sit between the two: random removal of a few mid layers is also
tolerated, consistent with partial pruning tolerance in the middle of the stack, but only the cosine order
keeps error flat out to eight layers. Beyond $k\!\approx\!12$ all orders degrade and
the curves are non-monotonic in $k$, because identity-patching interacts non-linearly
across the residual stream; we therefore read the frontier only in the clean
$k\!\le\!8$ regime. Planning-token input-output cosine, rather than marginal
decodability, is the better predictor of which layers this model can shed.

\section{Pruning Cost by Scenario}
\label{sec:robust}

\begin{figure}[tbp]
  \centering
  \includegraphics[width=0.72\textwidth]{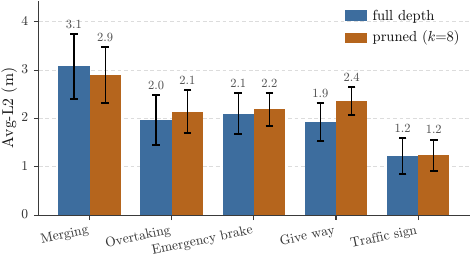}
  \caption{Per-family Avg-L2 for full depth and $k=8$ on the same 160 frames per
  family. Bars show marginal 95\% normal-approximation intervals. No family-specific
  difference is statistically resolved; the plot does not establish equivalence.}
  \label{fig:robust}
\end{figure}

\Cref{fig:robust} compares full-depth and $k=8$ predictions on the same 160
cold-start frames per family. The plotted intervals are marginal
normal-approximation intervals, $1.96\sigma/\sqrt{n}$. No family-specific
degradation is statistically resolved at this sample size (for example, traffic
sign $1.22\to1.24$\,m, emergency brake $2.10\to2.19$\,m, merging $3.08\to2.90$\,m;
$n=160$ each). However, overlapping marginal intervals are not an equivalence test
and cannot exclude small effects.

\section{Present Early, Not Yet Planner-Ready}
\label{sec:reshape}

\begin{figure}[tbp]
  \centering
  \includegraphics[width=0.85\textwidth]{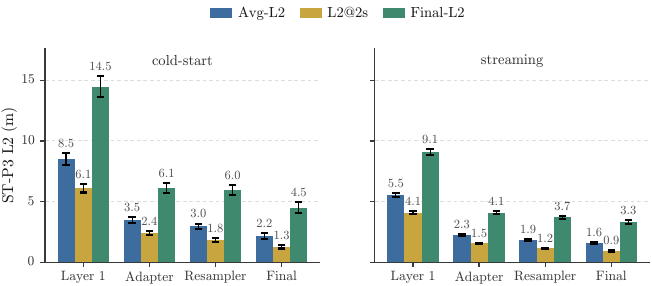}
  \caption{A learned readout of layer-1 representations. A residual adapter (acting on the
  planning token) and a sequence resampler (acting on the full sequence) both recover much of
  the layer-1-to-final gap, the resampler more so, but neither reaches the final layer. Three
  ST-P3 metrics, cold-start and streaming, with 95\% CIs.}
  \label{fig:reshape}
\end{figure}

Our readout decodes each layer's token through the \emph{frozen} planner, which is
trained on final-layer tokens. This preserves the released planner and its final-layer
readout interface, but is conservative for intermediate hidden states: an early-layer
token may carry the information in a
form the frozen planner cannot use, exactly the concern that motivates the tuned
lens over the plain logit lens~\cite{belrose2023tunedlens}. We therefore investigate
whether small learned transforms can read more out of the layer-1 hidden states.
We train two separate heads to map the layer-1 representations to the final-layer token space
with the planner frozen (\cref{fig:reshape}; details in \cref{tab:hparams}).

A residual MLP adapter ($4096\to1024\to4096$ variables) trained via MSE minimization
on 3,536 disjoint training scenes maps the layer-1 planning token to lower the
Avg-L2 error from 8.51 to 3.47\,m. Concurrently, a sequence-reading cross-attention
resampler~\cite{jaegle2021perceiver} featuring 8 latent queries and a 1024-dimension
bottleneck ($\sim$40M parameters) is trained on the same scenes and attends strictly
to the full layer-1 sequence to reach 2.98\,m against the final-layer baseline of 2.17\,m.
Thus, much of the layer-1-to-final gap is recoverable by learned readouts. This
confirms that the native-planner curve measures representation compatibility rather
than the first presence of planning information.

To assess which input groups the full-sequence-trained resampler depends on,
we evaluate test-time input masking in \cref{tab:tokens}. \Cref{tab:tokens} is a test-time masking analysis of the single
resampler trained on the full sequence; the masked subsets are not separately retrained models.
Retaining only the 529 vision tokens stays within 0.15\,m of the unablated sequence,
whereas retaining only text or the planning token degrades to near the raw layer-1 baseline.
For the full-sequence-trained resampler, retaining vision tokens preserves most of the performance,
while masking them causes substantial degradation. Because these are test-time masks, the result
reflects both information removal and possible input-distribution shift and does not establish that
the omitted token subsets contain no recoverable information.

\begin{table}[!ht]
  \centering
  \caption{Training configurations for the auxiliary reshaping heads.}
  \label{tab:hparams}
  \small
  \begin{tabular}{ll}
    \toprule
    Setting & Value \\
    \midrule
    Trainable parameters & Head only \\
    Frozen components     & LLM baseline, final normalization, generative planner \\
    Optimizer            & AdamW, learning rate $2\times10^{-4}$ \\
    Weight decay         & $0.05$ \\
    Learning schedule    & Cosine decay, 200-step linear warmup \\
    Batch size           & 8 \\
    MLP Adapter scale    & $\sim$1\,M parameters \\
    Resampler scale      & $\sim$40\,M parameters \\
    \bottomrule
  \end{tabular}
\end{table}

\begin{table}[!ht]
  \centering
  \caption{Test-time token masking for the resampler trained on the full layer-1
  sequence. No subset-specific resampler is retrained, so results may include
  distribution shift from masking. ST-P3 L2 displacement values in meters.}
  \label{tab:tokens}
  \small
  \begin{tabular}{lcccc}
    \toprule
    Input subset & Token count & Avg-L2 & L2@2s & Final-L2 \\
    \midrule
    Full sequence            & 599 & 2.98 & 1.83 & 5.95 \\
    Vision tokens retained   & 529 & 3.13 & 1.97 & 6.10 \\
    Text tokens retained     & 70  & 5.73 & 4.14 & 9.65 \\
    Planning token retained  & 1   & 5.92 & 4.23 & 10.15 \\
    Top-64 by norm           & 64  & 5.78 & 4.09 & 9.98 \\
    Bottom-64 by norm        & 64  & 6.05 & 4.28 & 10.47 \\
    \bottomrule
  \end{tabular}
\end{table}

\section{Discussion}
\label{sec:discussion}

For the evaluated ORION checkpoint, increasing decoder depth progressively
improves compatibility with the frozen planner, while high-level command
information is linearly decodable much earlier. Our layer-pruning results
quantify this structural trade-off. At the selected
efficiency-accuracy threshold, eliminating a quarter of the decoder layers incurs a
marginal increase in open-loop error without a statistically resolved family-specific
degradation at the evaluated sample size. However, this compression yields only a
1.33$\times$ decoder speedup. Because the decoder accounts for roughly half of the
per-frame latency, while perception components dominate the remaining budget, an
Amdahl-style projection based on the measured decoder share gives an end-to-end speedup
of approximately 1.13$\times$; this value was not directly timed end to end.
More substantial layer removal causes severe open-loop trajectory degradation.
While these observations are bound to a single architecture and simulation benchmark,
they clarify the empirical limits of post-hoc structural compression.

\section{Limitations}
\label{sec:limits}

This study is limited to one ORION checkpoint and Bench2Drive evaluation setup. All
pruning results are open-loop and do not establish closed-loop safety or deployment
readiness. The command probe uses frame-level rather than route-grouped folds and may
benefit from route correlations; its standardizer is also fitted before fold
construction. The frozen planner was trained on final-layer representations, so
intermediate decoding measures native-planner compatibility rather than all
recoverable information. \Cref{tab:tokens} uses test-time masking and may include
distribution shift. Per-family pruning intervals are underpowered for equivalence
claims. Finally, pruning is performed without recalibration or fine-tuning, and the
projected end-to-end gain is limited because perception accounts for most of the
remaining latency.

\section{Conclusion}
\label{sec:conclusion}

We introduced a trajectory-space native-head lens for examining the planning token of
an ORION-based driving VLA. Navigation command information is linearly decodable after
the first decoder block, while compatibility with the frozen trajectory planner improves
across depth and reaches its lowest error at the final hidden state. Learned layer-1
readouts recover much of this gap, showing that early planning information is present but
not yet represented in the planner's expected format. For this checkpoint, ranking layers
by planning-token input-output cosine permits removal of 8 of 32 layers within an
approximately 5\% relative open-loop error increase and yields a measured 1.33$\times$
decoder speedup. Isolated intermediate decode quality is a poor pruning criterion because
layers that decode badly alone can remain important for subsequent computation. These
findings are diagnostic and checkpoint-specific and do not establish closed-loop safety.

\clearpage
\bibliographystyle{splncs04}
\bibliography{refs}

@inproceedings{fu2025orion,
  title     = {{ORION}: A Holistic End-to-End Autonomous Driving Framework by Vision-Language Instructed Action Generation},
  author    = {Fu, Haoyu and Zhang, Diankun and Zhao, Zongchuang and Cui, Jianfeng and Liang, Dingkang and Zhang, Chong and Zhang, Dingyuan and Xie, Hongwei and Wang, Bing and Bai, Xiang},
  booktitle = {Proceedings of the {IEEE/CVF} International Conference on Computer Vision ({ICCV})},
  pages     = {24823--24834},
  year      = {2025}
}

@inproceedings{tian2024drivevlm,
  title     = {{DriveVLM}: The Convergence of Autonomous Driving and Large Vision-Language Models},
  author    = {Tian, Xiaoyu and Gu, Junru and Li, Bailin and Liu, Yicheng and Wang, Yang and Zhao, Zhiyong and Zhan, Kun and Jia, Peng and Lang, Xianpeng and Zhao, Hang},
  booktitle = {Proceedings of the 8th Conference on Robot Learning ({CoRL})},
  series    = {Proceedings of Machine Learning Research},
  volume    = {270},
  pages     = {4698--4726},
  year      = {2024}
}

@inproceedings{shao2024lmdrive,
  title     = {{LMDrive}: Closed-Loop End-to-End Driving with Large Language Models},
  author    = {Shao, Hao and Hu, Yuxuan and Wang, Letian and Song, Guanglu and Waslander, Steven L. and Liu, Yu and Li, Hongsheng},
  booktitle = {Proceedings of the {IEEE/CVF} Conference on Computer Vision and Pattern Recognition ({CVPR})},
  year      = {2024}
}

@inproceedings{sima2024drivelm,
  title     = {{DriveLM}: Driving with Graph Visual Question Answering},
  author    = {Sima, Chonghao and Renz, Katrin and Chitta, Kashyap and Chen, Li and Zhang, Hanxue and Xie, Chengen and Bei{\ss}wenger, Jens and Luo, Ping and Geiger, Andreas and Li, Hongyang},
  booktitle = {Proceedings of the European Conference on Computer Vision ({ECCV})},
  pages     = {256--274},
  year      = {2024}
}

@inproceedings{mao2023gptdriver,
  title     = {{GPT-Driver}: Learning to Drive with {GPT}},
  author    = {Mao, Jiageng and Qian, Yuxi and Ye, Junjie and Zhao, Hang and Wang, Yue},
  booktitle = {NeurIPS 2023 Workshop on Foundation Models for Decision Making},
  year      = {2023}
}

@article{xu2024drivegpt4,
  title   = {{DriveGPT4}: Interpretable End-to-End Autonomous Driving via Large Language Model},
  author  = {Xu, Zhenhua and Zhang, Yujia and Xie, Enze and Zhao, Zhen and Guo, Yong and Wong, Kwan-Yee K. and Li, Zhenguo and Zhao, Hengshuang},
  journal = {{IEEE} Robotics and Automation Letters},
  volume  = {9},
  number  = {10},
  pages   = {8186-8193},
  year    = {2024}
}

@article{hwang2024emma,
  title   = {{EMMA}: End-to-End Multimodal Model for Autonomous Driving},
  author  = {Hwang, Jyh-Jing and Xu, Runsheng and Lin, Hubert and Hung, Wei-Chih and Ji, Jingwei and Choi, Kristy and Huang, Di and He, Tong and Covington, Paul and Sapp, Benjamin and Zhou, Yin and Guo, James and Anguelov, Dragomir and Tan, Mingxing},
  journal = {Transactions on Machine Learning Research ({TMLR})},
  year    = {2025}
}

@misc{jiang2024senna,
  title         = {{Senna}: Bridging Large Vision-Language Models and End-to-End Autonomous Driving},
  author        = {Jiang, Bo and Chen, Shaoyu and Liao, Bencheng and Zhang, Xingyu and Yin, Wei and Zhang, Qian and Huang, Chang and Liu, Wenyu and Wang, Xinggang},
  year          = {2024},
  eprint        = {2410.22313},
  archivePrefix = {arXiv},
  primaryClass  = {cs.CV}
}

@misc{wang2023drivemlm,
  title         = {{DriveMLM}: Aligning Multi-Modal Large Language Models with Behavioral Planning States for Autonomous Driving},
  author        = {Wang, Wenhai and Xie, Jiangwei and Hu, ChuanYang and Zou, Haoming and Fan, Jianan and Tong, Wenwen and Wen, Yang and Wu, Silei and Deng, Hanming and Li, Zhiqi and Tian, Hao and Lu, Lewei and Zhu, Xizhou and Wang, Xiaogang and Qiao, Yu and Dai, Jifeng},
  year          = {2023},
  eprint        = {2312.09245},
  archivePrefix = {arXiv},
  primaryClass  = {cs.CV}
}

@inproceedings{mao2024agentdriver,
  title     = {A Language Agent for Autonomous Driving},
  author    = {Mao, Jiageng and Ye, Junjie and Qian, Yuxi and Pavone, Marco and Wang, Yue},
  booktitle = {Conference on Language Modeling ({COLM})},
  year      = {2024}
}

@inproceedings{nie2024reason2drive,
  title     = {{Reason2Drive}: Towards Interpretable and Chain-Based Reasoning for Autonomous Driving},
  author    = {Nie, Ming and Peng, Renyuan and Wang, Chunwei and Cai, Xinyue and Han, Jianhua and Xu, Hang and Zhang, Li},
  booktitle = {Proceedings of the European Conference on Computer Vision ({ECCV})},
  pages     = {292--308},
  year      = {2024}
}

@inproceedings{zheng2024genad,
  title     = {{GenAD}: Generative End-to-End Autonomous Driving},
  author    = {Zheng, Wenzhao and Song, Ruiqi and Guo, Xianda and Zhang, Chenming and Chen, Long},
  booktitle = {Proceedings of the European Conference on Computer Vision ({ECCV})},
  pages     = {87--104},
  year      = {2024}
}

@inproceedings{yang2024genad,
  title     = {{GenAD}: Generalized Predictive Model for Autonomous Driving},
  author    = {Yang, Jiazhi and Gao, Shenyuan and Qiu, Yihang and Chen, Li and Li, Tianyu and Dai, Bo and Chitta, Kashyap and Wu, Penghao and Zeng, Jia and Luo, Ping and Zhang, Jun and Geiger, Andreas and Qiao, Yu and Li, Hongyang},
  booktitle = {Proceedings of the {IEEE/CVF} Conference on Computer Vision and Pattern Recognition ({CVPR})},
  year      = {2024}
}

@inproceedings{kim2024openvla,
  title     = {{OpenVLA}: An Open-Source Vision-Language-Action Model},
  author    = {Kim, Moo Jin and Pertsch, Karl and Karamcheti, Siddharth and Xiao, Ted and Balakrishna, Ashwin and Nair, Suraj and Rafailov, Rafael and Foster, Ethan and Lam, Grace and Sanketi, Pannag R. and Vuong, Quan and Kollar, Thomas and Burchfiel, Benjamin and Tedrake, Russ and Sadigh, Dorsa and Levine, Sergey and Liang, Percy and Finn, Chelsea},
  booktitle = {Conference on Robot Learning ({CoRL})},
  year      = {2024}
}

@inproceedings{zitkovich2023rt2,
  title     = {{RT-2}: Vision-Language-Action Models Transfer Web Knowledge to Robotic Control},
  author    = {Zitkovich, Brianna and Yu, Tianhe and Xu, Sichun and Xu, Peng and Xiao, Ted and Xia, Fei and Wu, Jialin and Wohlhart, Paul and Welker, Stefan and Wahid, Ayzaan and Brohan, Anthony and others},
  booktitle = {Proceedings of the 7th Conference on Robot Learning ({CoRL})},
  series    = {Proceedings of Machine Learning Research},
  volume    = {229},
  pages     = {2165--2183},
  year      = {2023}
}

@inproceedings{jaegle2021perceiver,
  title     = {Perceiver: General Perception with Iterative Attention},
  author    = {Jaegle, Andrew and Gimeno, Felix and Brock, Andy and Vinyals, Oriol and Zisserman, Andrew and Carreira, Jo{\~a}o},
  booktitle = {Proceedings of the International Conference on Machine Learning ({ICML})},
  series    = {Proceedings of Machine Learning Research},
  volume    = {139},
  pages     = {4651--4664},
  year      = {2021}
}

@inproceedings{vaswani2017attention,
  title     = {Attention Is All You Need},
  author    = {Vaswani, Ashish and Shazeer, Noam and Parmar, Niki and Uszkoreit, Jakob and Jones, Llion and Gomez, Aidan N. and Kaiser, Lukasz and Polosukhin, Illia},
  booktitle = {Advances in Neural Information Processing Systems ({NeurIPS})},
  year      = {2017}
}

@misc{touvron2023llama,
  title         = {{LLaMA}: Open and Efficient Foundation Language Models},
  author        = {Touvron, Hugo and Lavril, Thibaut and Izacard, Gautier and Martinet, Xavier and Lachaux, Marie-Anne and Lacroix, Timoth{\'e}e and Rozi{\`e}re, Baptiste and Goyal, Naman and Hambro, Eric and Azhar, Faisal and Rodriguez, Aurelien and Joulin, Armand and Grave, Edouard and Lample, Guillaume},
  year          = {2023},
  eprint        = {2302.13971},
  archivePrefix = {arXiv},
  primaryClass  = {cs.CL}
}

@inproceedings{hu2023uniad,
  title     = {Planning-Oriented Autonomous Driving},
  author    = {Hu, Yihan and Yang, Jiazhi and Chen, Li and Li, Keyu and Sima, Chonghao and Zhu, Xizhou and Chai, Siqi and Du, Senyao and Lin, Tianwei and Wang, Wenhai and Lu, Lewei and Jia, Xiaosong and Liu, Qiang and Dai, Jifeng and Qiao, Yu and Li, Hongyang},
  booktitle = {Proceedings of the {IEEE/CVF} Conference on Computer Vision and Pattern Recognition ({CVPR})},
  pages     = {17853--17862},
  year      = {2023}
}

@inproceedings{jiang2023vad,
  title     = {{VAD}: Vectorized Scene Representation for Efficient Autonomous Driving},
  author    = {Jiang, Bo and Chen, Shaoyu and Xu, Qing and Liao, Bencheng and Chen, Jiajie and Zhou, Helong and Zhang, Qian and Liu, Wenyu and Huang, Chang and Wang, Xinggang},
  booktitle = {Proceedings of the {IEEE/CVF} International Conference on Computer Vision ({ICCV})},
  year      = {2023}
}

@inproceedings{hu2022stp3,
  title     = {{ST-P3}: End-to-End Vision-Based Autonomous Driving via Spatial-Temporal Feature Learning},
  author    = {Hu, Shengchao and Chen, Li and Wu, Penghao and Li, Hongyang and Yan, Junchi and Tao, Dacheng},
  booktitle = {Proceedings of the European Conference on Computer Vision ({ECCV})},
  year      = {2022}
}

@article{chitta2023transfuser,
  title   = {{TransFuser}: Imitation with Transformer-Based Sensor Fusion for Autonomous Driving},
  author  = {Chitta, Kashyap and Prakash, Aditya and Jaeger, Bernhard and Yu, Zehao and Renz, Katrin and Geiger, Andreas},
  journal = {IEEE Transactions on Pattern Analysis and Machine Intelligence},
  volume  = {45},
  number  = {11},
  pages   = {12878--12895},
  year    = {2023}
}

@inproceedings{prakash2021multimodal,
  title     = {Multi-Modal Fusion Transformer for End-to-End Autonomous Driving},
  author    = {Prakash, Aditya and Chitta, Kashyap and Geiger, Andreas},
  booktitle = {Proceedings of the {IEEE/CVF} Conference on Computer Vision and Pattern Recognition ({CVPR})},
  year      = {2021}
}

@inproceedings{alain2017probes,
  title     = {Understanding Intermediate Layers Using Linear Classifier Probes},
  author    = {Alain, Guillaume and Bengio, Yoshua},
  booktitle = {International Conference on Learning Representations ({ICLR}), Workshop Track},
  year      = {2017}
}

@misc{nostalgebraist2020logitlens,
  title        = {interpreting {GPT}: the logit lens},
  author       = {{nostalgebraist}},
  year         = {2020},
  howpublished = {LessWrong blog post},
  note         = {\url{https://www.lesswrong.com/posts/AcKRB8wDpdaN6v6ru/interpreting-gpt-the-logit-lens}}
}

@misc{belrose2023tunedlens,
  title         = {Eliciting Latent Predictions from Transformers with the Tuned Lens},
  author        = {Belrose, Nora and Furman, Zach and Smith, Logan and Halawi, Danny and Ostrovsky, Igor and McKinney, Lev and Biderman, Stella and Steinhardt, Jacob},
  year          = {2023},
  eprint        = {2303.08112},
  archivePrefix = {arXiv},
  primaryClass  = {cs.LG}
}

@inproceedings{fan2020layerdrop,
  title     = {Reducing Transformer Depth on Demand with Structured Dropout},
  author    = {Fan, Angela and Grave, Edouard and Joulin, Armand},
  booktitle = {International Conference on Learning Representations ({ICLR})},
  year      = {2020}
}

@article{sajjad2023dropping,
  title   = {On the Effect of Dropping Layers of Pre-trained Transformer Models},
  author  = {Sajjad, Hassan and Dalvi, Fahim and Durrani, Nadir and Nakov, Preslav},
  journal = {Computer Speech \& Language},
  volume  = {77},
  pages   = {101429},
  year    = {2023}
}

@inproceedings{men2025shortgpt,
  title     = {{ShortGPT}: Layers in Large Language Models are More Redundant Than You Expect},
  author    = {Men, Xin and Xu, Mingyu and Zhang, Qingyu and Yuan, Qianhao and Wang, Bingning and Lin, Hongyu and Lu, Yaojie and Han, Xianpei and Chen, Weipeng},
  booktitle = {Findings of the Association for Computational Linguistics: {ACL} 2025},
  pages     = {20192--20204},
  year      = {2025}
}

@inproceedings{gromov2025unreasonable,
  title     = {The Unreasonable Ineffectiveness of the Deeper Layers},
  author    = {Gromov, Andrey and Tirumala, Kushal and Shapourian, Hassan and Glorioso, Paolo and Roberts, Daniel A.},
  booktitle = {International Conference on Learning Representations ({ICLR})},
  year      = {2025}
}

@inproceedings{ma2023llmpruner,
  title     = {{LLM-Pruner}: On the Structural Pruning of Large Language Models},
  author    = {Ma, Xinyin and Fang, Gongfan and Wang, Xinchao},
  booktitle = {Advances in Neural Information Processing Systems ({NeurIPS})},
  year      = {2023}
}

@inproceedings{ashkboos2024slicegpt,
  title     = {{SliceGPT}: Compress Large Language Models by Deleting Rows and Columns},
  author    = {Ashkboos, Saleh and Croci, Maximilian L. and do Nascimento, Marcelo Gennari and Hoefler, Torsten and Hensman, James},
  booktitle = {International Conference on Learning Representations ({ICLR})},
  year      = {2024}
}

@inproceedings{elhoushi2024layerskip,
  title     = {{LayerSkip}: Enabling Early Exit Inference and Self-Speculative Decoding},
  author    = {Elhoushi, Mostafa and Shrivastava, Akshat and Liskovich, Diana and Hosmer, Basil and Wasti, Bram and Lai, Liangzhen and Mahmoud, Anas and Acun, Bilge and Agarwal, Saurabh and Roman, Ahmed and Aly, Ahmed A. and Chen, Beidi and Wu, Carole-Jean},
  booktitle = {Proceedings of the Annual Meeting of the Association for Computational Linguistics ({ACL})},
  year      = {2024}
}

@inproceedings{teerapittayanon2016branchynet,
  title     = {{BranchyNet}: Fast Inference via Early Exiting from Deep Neural Networks},
  author    = {Teerapittayanon, Surat and McDanel, Bradley and Kung, H. T.},
  booktitle = {Proceedings of the International Conference on Pattern Recognition ({ICPR})},
  year      = {2016}
}

@inproceedings{schuster2022confident,
  title     = {Confident Adaptive Language Modeling},
  author    = {Schuster, Tal and Fisch, Adam and Gupta, Jai and Dehghani, Mostafa and Bahri, Dara and Tran, Vinh Q. and Tay, Yi and Metzler, Donald},
  booktitle = {Advances in Neural Information Processing Systems ({NeurIPS})},
  year      = {2022}
}

@inproceedings{jia2024bench2drive,
  title     = {Bench2Drive: Towards Multi-Ability Benchmarking of Closed-Loop End-To-End Autonomous Driving},
  author    = {Jia, Xiaosong and Yang, Zhenjie and Li, Qifeng and Zhang, Zhiyuan and Yan, Junchi},
  booktitle = {Advances in Neural Information Processing Systems ({NeurIPS}) Datasets and Benchmarks Track},
  year      = {2024}
}

@inproceedings{dosovitskiy2017carla,
  title     = {{CARLA}: An Open Urban Driving Simulator},
  author    = {Dosovitskiy, Alexey and Ros, German and Codevilla, Felipe and Lopez, Antonio and Koltun, Vladlen},
  booktitle = {Proceedings of the Conference on Robot Learning ({CoRL})},
  year      = {2017}
}

\end{document}